\documentclass[10pt,twocolumn,letterpaper]{article}

\usepackage{cvpr}              
\usepackage{bbding}
\usepackage{multirow}
\usepackage{algorithm}
\usepackage{algpseudocode}
\usepackage{makecell}

\definecolor{cvprblue}{rgb}{0.21,0.49,0.74}
\usepackage[pagebackref,breaklinks,colorlinks,citecolor=cvprblue]{hyperref}
\usepackage[misc]{ifsym} 

\def\paperID{} 
\def\confName{CVPR}
\def\confYear{2026}

\title{SceneScribe-1M: A Large-Scale Video Dataset with Comprehensive \\  Geometric and Semantic Annotations}

\author{Yunnan Wang\textsuperscript{1,2,4,5*}, Kecheng Zheng\textsuperscript{2*}, Jianyuan Wang\textsuperscript{3}, Minghao Chen\textsuperscript{3}, David Novotny, \\ Christian Rupprecht\textsuperscript{3},  Yinghao Xu\textsuperscript{2}, Xing Zhu\textsuperscript{2}, Wenjun Zeng\textsuperscript{4,5}, Xin Jin\textsuperscript{4,5}, Yujun Shen\textsuperscript{2 \text{\Letter}}\\
\textsuperscript{1} Shanghai Jiao Tong University \quad \textsuperscript{2} Ant Group\quad \textsuperscript{3} Visual Geometry Group, University of Oxford\\
 \textsuperscript{4} Ningbo Institute of Digital Twin, Eastern Institute of Technology, Ningbo \\ \textsuperscript{5} Zhejiang Key Laboratory of Industrial Intelligence and Digital Twin \\
\rmfamily\href{https://wangyunnan.github.io/SceneScribe-1M}{https://wangyunnan.github.io/SceneScribe-1M}
}

\begin{document}
\twocolumn[{
\renewcommand\twocolumn[1][]{#1}%
\maketitle
\vspace{-28pt}
\begin{center}
\includegraphics[width=0.99\textwidth]{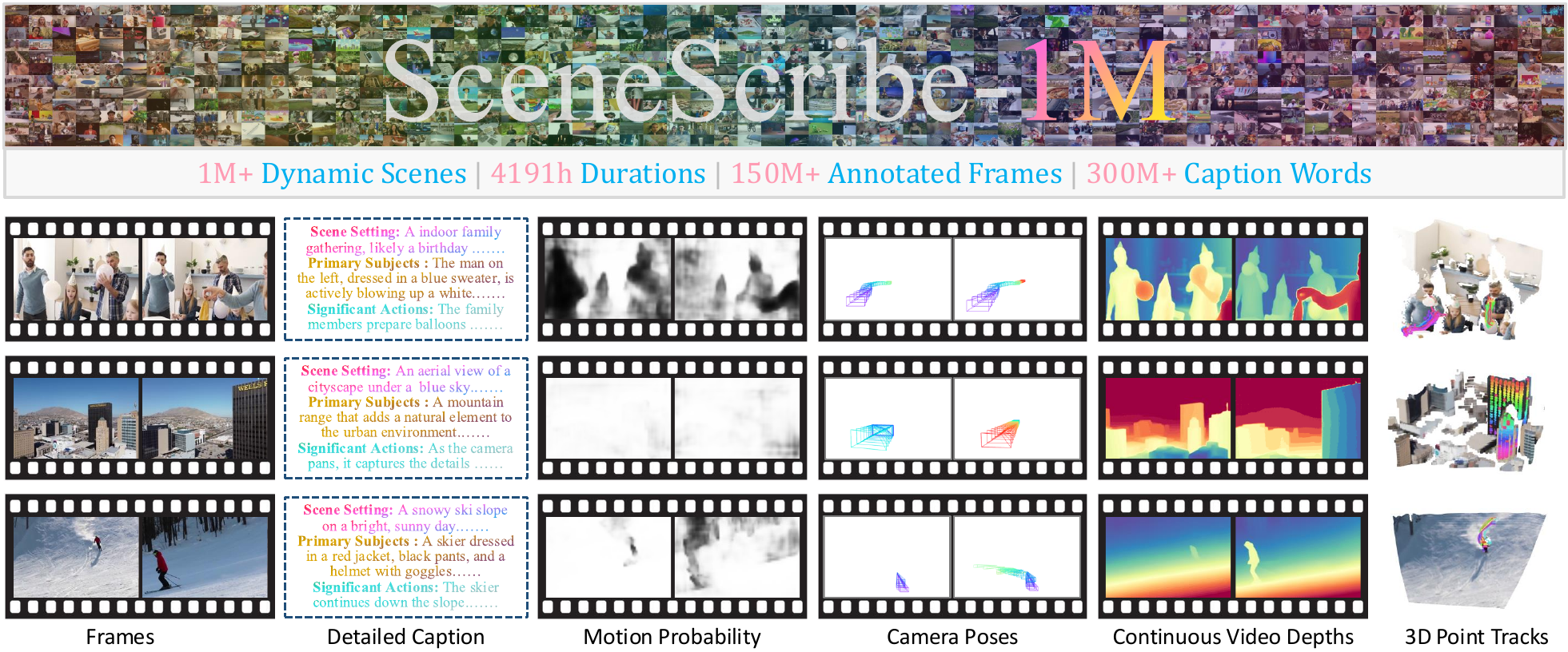}
\end{center}
\vspace{-0.5cm}
\captionof{figure}{\textbf{SceneScribe-1M} offers more than one million dynamic scenes spanning over 4,000 hours, featuring \textit{comprehensive semantic and geometric annotations} (\textit{i.e.}, detailed description, motion masks, camera poses, continuous video depths, and dynamic tracks). It supports \textit{diverse downstream tasks} (\textit{i.e.}, modular depth estimation, scene reconstruction, dynamic point tracking, and pose/text-to-video generation).}
\label{teaser}
\vspace{0.6cm}
}]
\def\thefootnote{*}\footnotetext{Equal Contribution. \ \textsuperscript{\text{\Letter}} Corresponding author.}
\def\thefootnote{\arabic{footnote}} 

\begin{abstract}
The convergence of 3D geometric perception and video synthesis has created an unprecedented demand for large-scale video data that is rich in both semantic and spatio-temporal information. While existing datasets have advanced either 3D understanding or video generation, a significant gap remains in providing a unified resource that supports both domains at scale. To bridge this chasm, we introduce SceneScribe-1M, a new large-scale, multi-modal video dataset. It comprises one million in-the-wild videos, each meticulously annotated with detailed textual descriptions, precise camera parameters, dense depth maps, and consistent 3D point tracks. We demonstrate the versatility and value of SceneScribe-1M by establishing benchmarks across a wide array of downstream tasks, including monocular depth estimation, scene reconstruction, and dynamic point tracking, as well as generative tasks such as text-to-video synthesis, with or without camera control. By open-sourcing SceneScribe-1M, we aim to provide a comprehensive benchmark and a catalyst for research, fostering the development of models that can both perceive the dynamic 3D world and generate controllable, realistic video content.
\end{abstract}    
\vspace{-0.85cm}
\section{Introduction}
\begin{table*}[!ht]
  \centering
  \caption{\textbf{Comparisons with Previous Works}. SceneScribe-1M is a large-scale video dataset with comprehensive geometric and semantic annotations. In the Geometric Annotation column, Depth map, Camera Pose, and 3D Tracks are abbreviated as D., C., and P., respectively.}
  \vspace{-0.2cm}
  \resizebox{\linewidth}{!}{
    \begin{tabular}{clcccccc}
    \toprule
    \multicolumn{1}{c}{\textbf{Type}} & \textbf{Dataset} & \textbf{Domain} & \textbf{Dynamic} & \textbf{Sem. Ann.} & \textbf{3D. Ann.} & \textbf{\#Scene Clips } & \textbf{\#Frames} \\
    \midrule
    \multirow{7}[2]{*}{3D Perception} & RealEstate10K~\cite{zhou2018stereo} & Indoor-Real & \XSolidBrush    & N/A & C.    & 80K   & 10M \\
    &BlendedMVS~\cite{yao2020blendedmvs} & Open-Synthetic & \XSolidBrush    & Single Label & D. C.  & 113   & 17K \\
    &CO3Dv2~\cite{reizenstein2021common} & Object-Real & \XSolidBrush    & Single Label & C.    & 19K   & 1.5M \\
    &PointOdyssey~\cite{zheng2023pointodyssey} & Object-Synthetic &  \CheckmarkBold    & N/A & D. C.  P. & 159   & 200K \\
    &CamVid-30K~\cite{zhao2024genxd} & Open-Real & \CheckmarkBold   & N/A & C.    & 30K   & - \\
    &Multi-Cam Video & Open-Synthetic & \CheckmarkBold   & Single Label & C.    & 136K  & 11M \\
    &DynPose-100K~\cite{rockwell2025dynamic} & Open-Real & \CheckmarkBold   & {Short Caption} & C.    & 100K  & 6.8M \\
    \midrule
    \multirow{3}[2]{*}{\makecell{Generation\\\& \\Understanding}} &HD-VILA-100M~\cite{xue2022advancing} & Open-Real & {\CheckmarkBold}   & Short Caption & N/A   & 103M  & 760k \\
    &Panda-70M~\cite{chen2024panda} & Open-Real & {\CheckmarkBold}   & Short Caption & N/A   & 70M   & 167K \\
    &Koala-36M~\cite{wang2025koala} & Open-Real & {\CheckmarkBold}   & Long Caption & N/A   & 36M   & 172k \\
    \midrule
    \multirow{3}[2]{*}{WFM}&Sekai-Real~\cite{li2025sekai} & Open-Real & {\CheckmarkBold}   & Structured Caption & D. C.  & $\sim$0.4M  & $\sim$40M \\
&SpatialVID~\cite{wang2025spatialvid} & Open-Real & {\CheckmarkBold}   & Structured Caption & D. C.  & $\sim$2M    & 123.6M \\
    &SceneScribe-1M (\textbf{ours}) & Open-Real & {\CheckmarkBold}   & Structured Caption & D. C.  P. & $\sim$1M    & 156.7M \\
    \bottomrule
    \end{tabular}}%
  \label{tab:source_video}%
  \vspace{-0.3cm}
\end{table*}%

\label{sec:intro}
In recent years, the rapid advancement of 3D geometric perception and video synthesis have significantly accelerated research in world foundation models (WFMs)~\cite{deepmind_genie3_2024,ali2025world,li2025sekai}. 
Collectively, these technologies enable WFMs to perceive, simulate, and interact effectively within dynamic environments.
Such capabilities integrated by WFMs are critical for promoting transformative developments in areas such as augmented reality~\cite{huang2025voyager}, robotics~\cite{fu2025learning,cen2025worldvla}, and autonomous driving~\cite{li2025omninwm,li2025drivevla}.
However, the scarcity of sufficiently large and high-quality datasets restricts the potential of existing models in both 3D perception and video synthesis, thereby further hindering the prospects of WFMs.

Current efforts to address data challenges related to 3D perception can be categorized into two main paradigms. 
One common strategy~\cite{cabon2020virtual,yao2020blendedmvs,bai2025recammaster} follows a data synthesis pipeline within virtual engines, automatically generating ground-truth camera poses and corresponding geometric annotations. 
Nevertheless, these approaches introduce a domain gap and overlook complex physical interactions.
Alternatively, another prevalent routine attempts to efficiently annotate real-world data by SfM~\cite{schonberger2016structure} or SLAM~\cite{mur2015orb} systems. Apart from the sparsity of camera trajectory annotations in static scenes~\cite{rockwell2025dynamic}, the annotation scale and diversity for dynamic scenes are also limited by computational overhead~\cite{zhou2018stereo,zhao2024genxd}.
Beyond 3D perception, video generation data with rich semantic information is also essential for building WFMs. Notably, current open-world datasets~\cite{nan2024openvid,wang2025koala,chen2024panda} have somewhat alleviated the issues of limited data and annotation scarcity present in previous studies~\cite{soomro2012ucf101,zhu2022celebv,yuan2024chronomagic}.
Nonetheless, since these datasets are tailored for video generation (e.g., text-to-video~\cite{kong2024hunyuanvideo}), they lack geometric annotations, consequently leaving the semantic and motion diversity required by WFMs insufficiently examined.
Despite the above progress of single-modal datasets, advances in WFMs remain fundamentally constrained by the inadequacy of large-scale datasets that comprehensively capture 3D geometric and fine-grained semantic properties.

In this paper, we introduce SceneScribe-1M, a large-scale, multi-modal video dataset that facilitates the critical intersection of 3D geometric perception and video synthesis (as shown in Figure~\ref{teaser}).
By incorporating powerful models in proprietary domains (\textit{i.e.}, Qwen2.5-VL-72B~\cite{bai2025qwen2}, MegaSaM~\cite{li2025megasam}, and TAPIP3D~\cite{zhang2025tapip3d}), we deploy over 1,000 GPUs to implement our annotation pipeline on large-scale videos.
SceneScribe-1M comprises one million in-the-wild videos, amounting to over 4,000 hours, each extensively annotated with detailed textual descriptions, precise camera parameters, continuous video depths, and consistent 3D point tracks.
Crucially, our curation establishes criteria across four key aspects, informed by both semantic and geometric annotations: video parameters, semantic information, camera motion, and object motion. Raw videos are meticulously examined based on these indicators to ensure content diversity and motion richness.
We further devise a filtering mechanism for SceneScribe-MVS subset construction, aiming to accommodate multi-view tasks that prefer static objects. 
This filter disentangles the camera and object motion, controlling the dynamic object inclusion without compromising camera motion intensity.
To establish rigorous benchmarks, we leverage SceneScribe-1M for core 3D perception, including monocular depth estimation, scene reconstruction, and dynamic point tracking. 
Moreover, SceneScribe-1M serves as a pivotal resource for advancing generative tasks such as text/pose-to-video synthesis, supporting precise view control over camera motion.


\begin{figure*}[!ht]
  \centering
    \includegraphics[width=\linewidth]{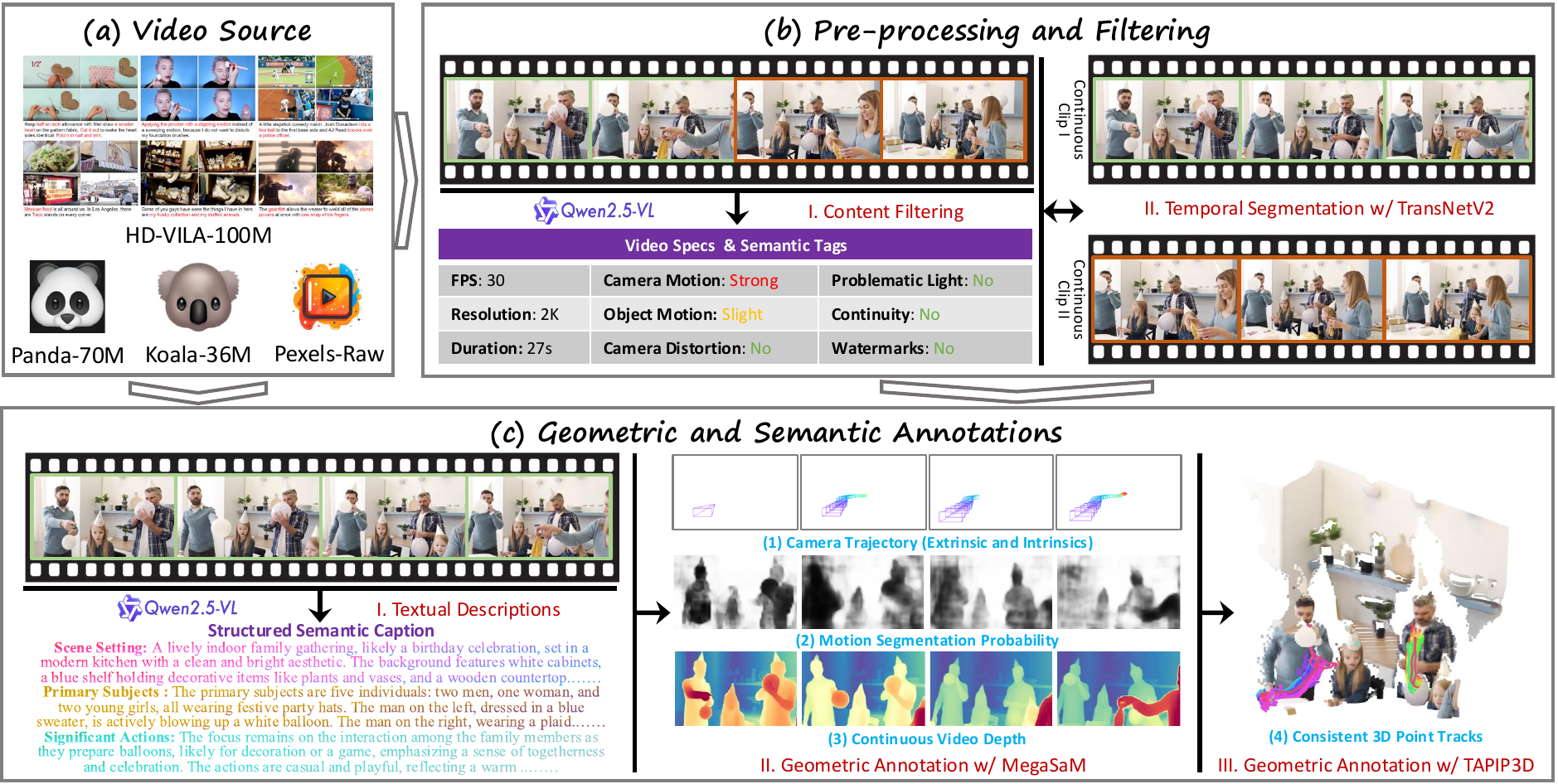}
      \vspace{-0.6cm}
  \caption{\textbf{Curation Pipeline} for SceneScribe-1M consist of: (a) We begin by collecting large-scale videos from various sources; (b) Raw videos undergo specification and content inspection, with temporal segmentation models employed to ensure continuity; and (c) We integrate Qwen2.5-VL-72B~\cite{bai2025qwen2}, MegaSaM~\cite{li2025megasam}, and TAPIP3D~\cite{zhang2025tapip3d} to perform comprehensive geometric and semantic annotations.}
  \vspace{-0.2cm}
  \label{fig:framework}
\end{figure*}

In summary, our primary contributions are as follows:
\begin{itemize}
    \item \textbf{Comprehensive Video Annotations}: SceneScribe-1M contains over 4,000 hours of video data, accompanied by essential geometric and semantic annotations. These annotations provide a unified resource that facilitates both large-scale 3D perception and video generative tasks.
    \item \textbf{Curated Videos with Semantic and Motion Diversity}: 
     SceneScribe-1M is curated with semantic and geometric indicators for content and motion diversity. 
    We also introduce a multi-view filter for SceneScribe-MVS to limit dynamic objects while preserving camera motion.
    \item \textbf{Extensive Downstream Evaluation}: The potential versatility of SceneScribe-1M is demonstrated by its applicability across diverse downstream tasks, including 3D geometric perception and video synthesis, which in turn highlight both the effectiveness and the quality of the dataset.
\end{itemize}


\section{Related Work}
\label{sec:formatting}

\textbf{World Foundation Models}.
As a significant advancement of spatial intelligence, world foundation models (WFMs)~\cite{brooks2024video,deepmind_genie3_2024,ali2025world,meng2024towards,kang2024far} involve the perception, simulation, and interaction within dynamic scenes. Given these properties, 3D geometric perception (covering depth estimation~\cite{piccinelli2024unidepth,yang2024depth,wang2025moge}, scene reconstruction~\cite{wang2025vggt,zhang2024monst3r, wang2025continuous, zhang2025flare,li2025megasam}, and dynamic point tracking~\cite{karaev2025cotracker3,xiao2024spatialtracker,xiao2025spatialtrackerv2}) and video generation (covering text-to-video~\cite{kong2024hunyuanvideo,wang2024scene,hong2022cogvideo,wan2025}, image-to-video~\cite{blattmann2023stable,xing2024dynamicrafter,xu2024easyanimate}, and pose-to-video~\cite{bahmani2025ac3d,he2024cameractrl,bai2025recammaster,bai2024syncammaster}) have emerged as fundamental technologies of WFMs.
This paper presents a unified resource that integrates spatio-temporal semantic and geometric information, advancing WFMs from separate video generation or 3D perception to interactive simulations within virtual environments.

\noindent\textbf{Video Data with Geometric/Semantic Annotations}.
Existing datasets~\cite{yao2020blendedmvs,cabon2020virtual,zhou2018stereo,bai2024syncammaster, zhao2024genxd, rockwell2025dynamic} for 3D perception primarily provide annotations such as depth maps, camera poses, and dynamic tracks, facilitating spatial tasks like depth estimation, scene reconstruction, and dynamic point tracking.
Meanwhile, text-to-video datasets typically consist of video collections with various scales, accompanied by either brief~\cite{soomro2012ucf101,xu_2016_CVPR,zhu2022celebv,yuan2024chronomagic} or detailed~\cite{xue2022advancing,chen2024panda,wang2025koala,nan2024openvid} descriptions.
Despite the availability of these datasets, they frequently lack a comprehensive resource capable of supporting large-scale advancements in both 3D understanding and video generation.
Notably, concurrent studies demonstrate an increasing trend toward integrating spatial geometry and semantic information. However, these works remain constrained either by the data scale (600+ hours of Sekai~\cite{li2025sekai} compared to our 4,000+ hours) or the comprehensive geometric annotations (the lack of consistent 3D point tracks in SpatialVID~\cite{wang2025spatialvid}). As summarized in Table~\ref{tab:source_video}, SceneScribe-1M features comprehensive geometric and semantic annotations for dynamic scenes, demonstrating superior scale and applicability compared to existing datasets.

\section{SceneScribe-1M Curation}

As depicted in Figure~\ref{fig:framework}, the curating pipeline for SceneScribe-1M consists of three key steps: collection, pre-processing, and annotation.
In the following sections, we describe each step in detail:
(\textbf{\romannumeral1}) the raw video source and the selection criteria  (Section~\ref{sec:s1});
(\textbf{\romannumeral2}) the pre-processing procedures, including quality filtering and temporal segmentation (Section~\ref{sec:s2});
(\textbf{\romannumeral3}) the multi-modal annotation pipeline, covering textual descriptions, precise camera parameters, dense depth maps, motion masks, and consistent 3D point tracks (Section~\ref{sec:s3});
(\textbf{\romannumeral4}) the sampling strategy for filtering a multi-view subset SceneScribe-MVS  (Section~\ref{sec:s4}).

\subsection{Source Video Collection}
\label{sec:s1}

\noindent\textbf{Video Source for SceneScribe-1M}.
To ensure the diversity and scale of SceneScribe-1M, we start by incorporating publicly available large-scale text-video paired datasets, \textit{i.e.}, HD-VILA-100M~\cite{xue2022advancing}, Panda-70M~\cite{chen2024panda}, and Koala-36M~\cite{wang2025koala}.
With initial quality screening and extensive validation in both understanding and generation tasks, these resources offer a robust foundation for SceneScribe-1M.
Specifically, each source contributes distinct strengths: HD-VILA-100M supplies large-scale videos covering diverse categories; Panda-70M provides extensive video-caption pairs with rich semantics; and Koala-36M brings precise temporal segmentation.
In Table~\ref{tab:source_video}, we summarize statistics of these datasets.
While these large-scale datasets provide substantial diversity, our assessment suggests they exhibit certain limitations in the motion varieties of both the camera and objects.
As a result, there is a sharp decrease in dataset scale after filtering for motion diversity.
To address this issue, we further curate the Pexels-Video dataset by sourcing videos from Pexels, a platform renowned for its extensive and diverse video resources.
In particular, we employ the OpenVideo~\cite{openvideo} toolbox to harvest a dataset of $668$k high-quality videos from the official Pexels website.

\begin{figure}[t]
  \centering
      \begin{minipage}[b]{0.12\textwidth}
    \centering
    \includegraphics[width=\textwidth]{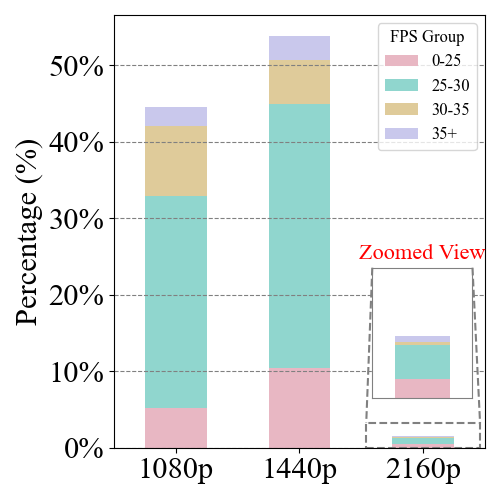}
    \centerline{(a) Resolution\&FPS}
    \label{fig:left9}
  \end{minipage}%
      \hspace{0.04\textwidth}
  \begin{minipage}[b]{0.24\textwidth}
    \centering
    \includegraphics[width=\textwidth]{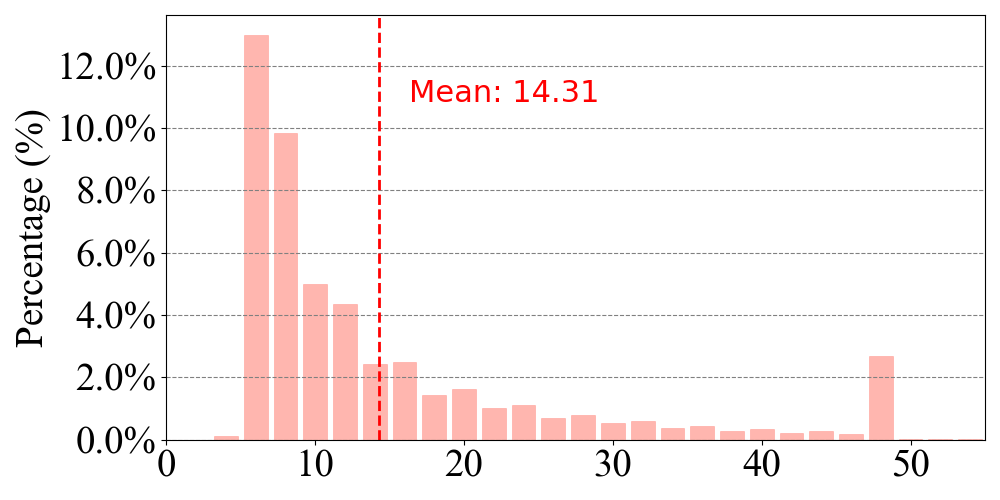}
    \centerline{(b) Duration (second)}
    \label{fig:left10}
  \end{minipage}%
  \vspace{-0.4cm}
  \caption{\textbf{Statistics of Raw Video Specification} after filtering, including Resolution, Frame Per Second (FPS), and Duration. }
   \vspace{-0.5cm}
    \label{fig:video_s}
\end{figure}

\begin{figure}[!t]
  \centering
  \includegraphics[width=0.45\textwidth]{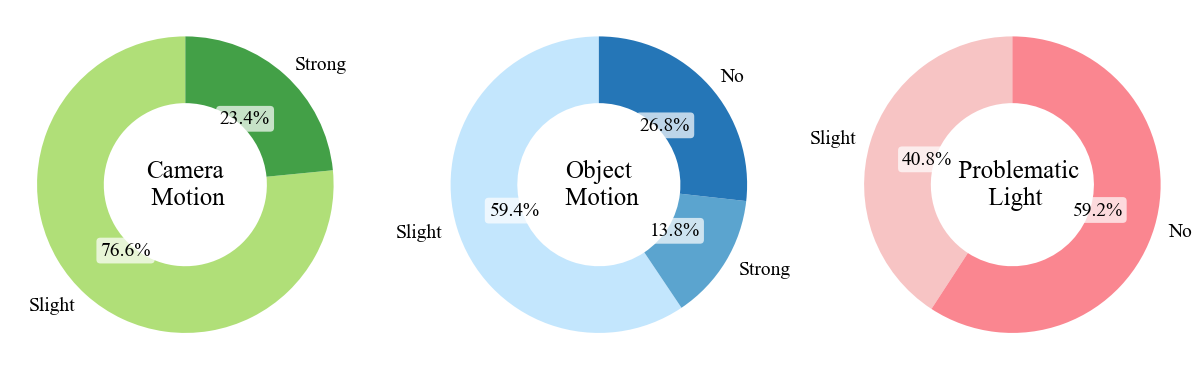}
   \vspace{-0.4cm}
  \caption{\textbf{Statistics of Raw Video Content} after filtering. These charts demonstrate that the raw videos exhibit sufficient diversity of motion while eliminating the lighting interference.}
    \vspace{-0.5cm}
    \label{fig:s_vc}
\end{figure}

%

\noindent\textbf{Selection Criteria}. 
To ensure precise annotation in SceneScribe-1M, we rigorously filter raw videos according to several criteria, including resolution, frame rate, and duration.
Specifically, we first select videos with spatial resolutions greater than 1080p to preserve fine-grained details.
Since low frame rates may hinder reliable motion detection and scene reconstruction, we prioritize videos with higher frame rates ($\geq10$ frames per second), which provide smoother transitions and enable accurate temporal alignment.
In addition, to facilitate comprehensive scene coverage, we opt for videos with durations spanning 5 seconds to 1 minute.
This is because shorter videos often lack sufficient scene variability, while longer videos substantially increase the costs of data processing and annotation.

\subsection{Video Pre-processing and Filtering}
\label{sec:s2}
\noindent\textbf{Quality Filtering}.
Despite an initial video screening by hard parameters, the content quality of the videos (\textit{e.g.}, camera perspective and object motion intensity) are not examined.
To optimize video suitability for both 3D geometric perception and video synthesis, we implement a comprehensive content filtering procedure, utilizing a powerful multimodal large language model (i.e., Qwen2.5-VL-72B~\cite{bai2025qwen2}) as an automated evaluator.
Specifically, we meticulously craft question templates across six critical dimensions to assess the source video content, as exemplified in Figure~\ref{fig:framework}.
Please refer to the \textbf{Supplementary Materials} for the detailed question templates.
Given these assessments, videos that fail to meet specific content quality thresholds, such as those exhibiting unknown motion intensity, visible watermarks, strong camera distortion,  or strong lighting artifacts, are excluded from the curated dataset. 

\begin{figure}[!t]
  \centering
  \begin{minipage}[b]{0.24\textwidth}
    \centering
    \includegraphics[width=\textwidth]{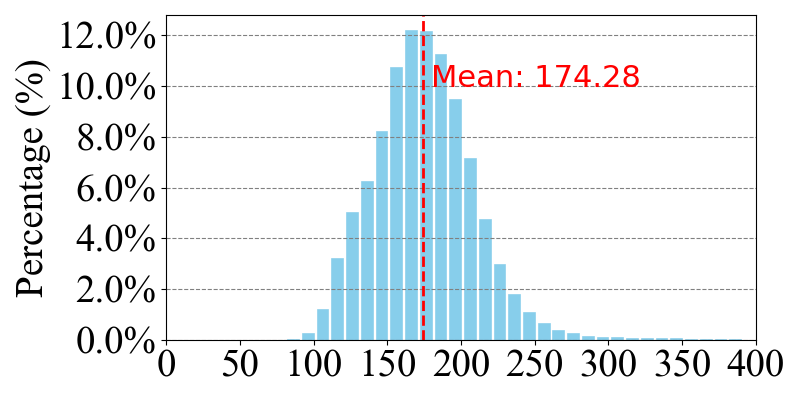}
    \centerline{(a) Caption Lengths (words)}
    \label{fig:left7}
  \end{minipage}%
  \begin{minipage}[b]{0.24\textwidth}
    \centering
    \includegraphics[width=\textwidth]{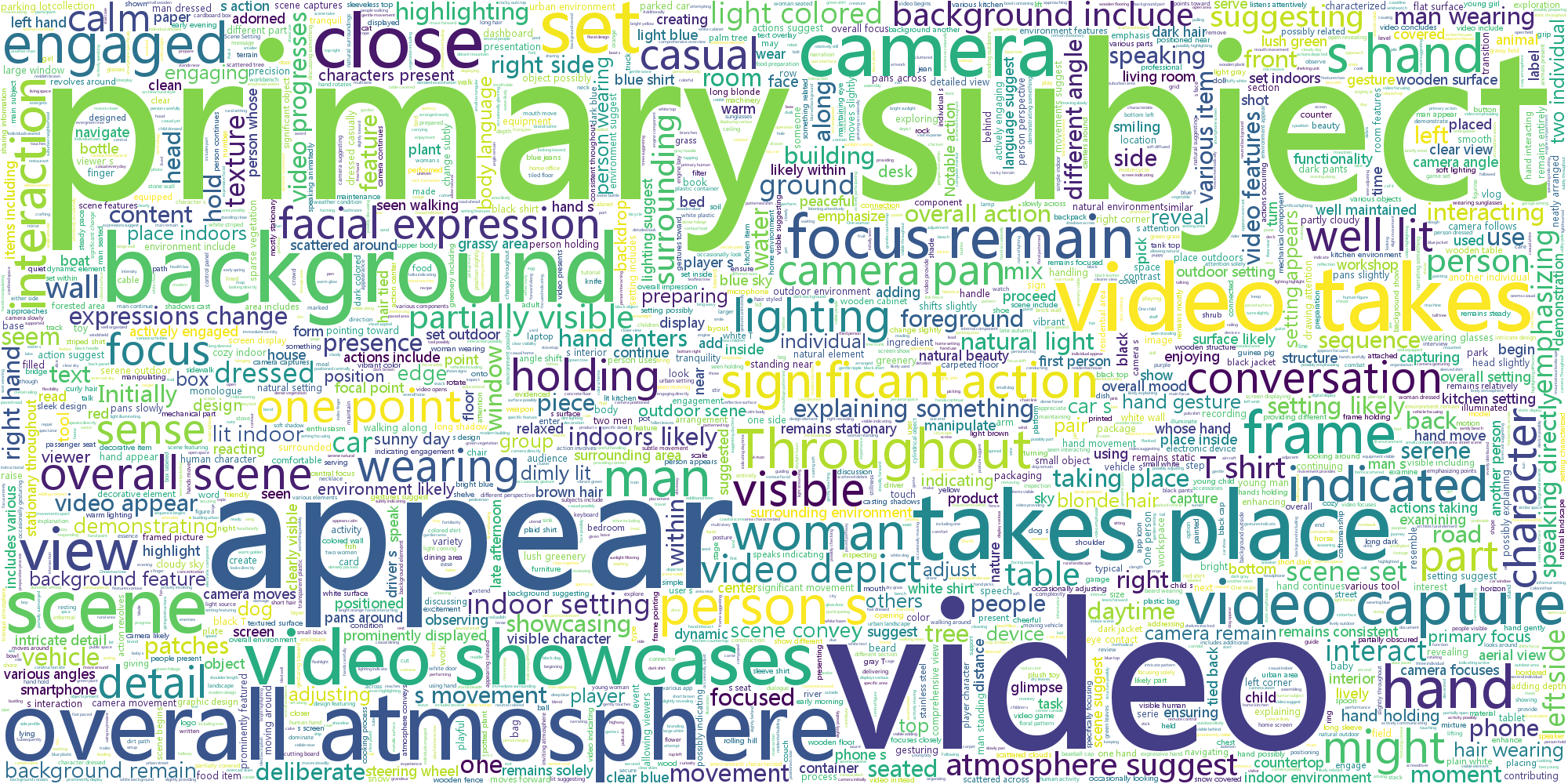}
    \centerline{(b) Word Cloud}
    \label{fig:right8}
  \end{minipage}
      \vspace{-0.8cm}
  \caption{\textbf{Caption Statistics}: (a) The average caption length is adequate to capture the details within each scene, and (b) Key words (\textit{e.g.,} \textit{atmosphere}, \textit{subject}, and \textit{take place}) effectively cover aspects such as the scene context, primary objects, and actions.}
  \label{fig:caption_s}
      \vspace{-0.1cm}
\end{figure}

\noindent\textbf{Temporal Segmentation for Non-Continuous Videos}.
Videos tagged as ``Non-Continuous" are inappropriate for both 3D vision (\textit{e.g.}, consistent 3D point tracking) and video generation.
Therefore, accurately partitioning these videos into temporal segments plays a vital role in dataset construction.
To achieve automatic and robust shot transition detection (\textit{e.g.}, hard cuts and gradual changes), we utilize TransNetV2~\cite{soucek2024transnet}, a model that achieves state-of-the-art results on respected benchmarks, enabling efficient processing of extensive video archives.
Effective segmentation along scene boundaries ensures that individual clips are semantically coherent, while these clips are subsequently re-filtered with the quality criteria. In Figure~\ref{fig:video_s} and~\ref{fig:s_vc}, we show the statistics of video parameters and content after filtering.

\subsection{Geometric and Semantic Annotation}
\label{sec:s3}
To facilitate comprehensive annotation of SceneSribe-1M, our pipeline integrates three distinct models, each optimized for a specific modality:
Qwen2.5-VL-72B~\cite{bai2025qwen2} for textual descriptions, MegaSaM~\cite{li2025megasam} for 3D geometric labeling, and TAPIP3D~\cite{zhang2025tapip3d} for dynamic point tracks. This multi-model framework guarantees extensive and high-quality annotation, thereby supporting diverse downstream applications in both 3D geometric perception and video synthesis.

\noindent\textbf{Semantic Annotation.}
We adopt Qwen2.5-VL-72B~\cite{bai2025qwen2} as the semantic annotation engine. 
Our choice is motivated by its performance, which is comparable to leading models such as GPT-4o~\cite{hurst2024gpt} and Gemini-2-Flash~\cite{deepmind2024gemini2} on various authoritative benchmarks, while excelling in visual understanding assessments.
By utilizing dynamic resolution processing and absolute temporal encoding, Qwen2.5-VL-72B is capable of handling long videos while precisely capturing events. This capability satisfies semantic requirements that demand extended temporal context and fine-grained action localization.
For each video, the model produces a comprehensive, structured scene description that clearly delineates scene settings, primary subjects or characters, and significant actions occurring.
Please refer to the \textbf{Supplementary Materials} for the detailed question templates.

\begin{algorithm}[!t]
    \caption{Multi-View Reprojection with Depth}
    \label{alg:reproj}
    \begin{algorithmic}[1] 
        \Require Reference depth $D_{r}$, Reference intrinsic $K_{r}$, Reference extrinsic $E_{r}$, Source depth $D_{s}$, Source Image $I_{s}$, Source intrinsic $K_{s}$, Source extrinsic $E_{s}$
        \Ensure Reprojected depth $D_{s2r}$, Reprojected image $I_{s2r}$, and Reprojected 2d coordinates $(x_{s2r}, y_{s2r})$
        \For{each pixel $(x_r, y_r)$ in $D_{r}$}
        \Statex {{\textit{Step \textcolor{red}{1}: Projecting 2D Points in Reference Pixel Coordinate to 3D Reference Camera Coordinate}}}
            \State $P_{r2c} \gets K_{r}^{-1} [x_r, y_r, 1]^T \cdot D_{r}(x_r, y_r)$
        \Statex {{\textit{Step \textcolor{red}{2}:  Projecting 3D Points in Reference Camera Coordinate to 2D Source Pixel Coordinate}}}
            \State $[P_{r2s};1] \gets E_{s} E_{r}^{-1} [P_{r2c};1]$
            \State $[u, v, w] \gets K_{s} \cdot P_{r2s}$
            \State $x_{r2s} \gets u/w$, $y_{r2s} \gets v/w$
        \Statex {{\textit{Step \textcolor{red}{3}: Sampling Source Depth Points and Projecting these Points to 3D Source Camera Coordinate}}}
            \State $I_{s2r} \gets I_{s}(x_{r2s}, y_{r2s})$
            \State $D'_{s} \gets D_{s}(x_{r2s}, y_{r2s})$
            \State $P_{s2c} \gets K_{s}^{-1} [x_{r2s}, y_{r2s}, 1]^T \cdot D'_{s}$
        \Statex {{\textit{Step \textcolor{red}{4}: Projecting 3D Points in Source Camera Coordinate to 2D Reference Pixel Coordinate}}}
            \State $[P_{s2r}; 1] \gets E_{r} E_{s}^{-1} [P_{s2c}; 1]$
            \State $[u', v', w'] \gets K_{r} \cdot P_{s2r}$
            \State $x_{s2r} \gets u'/w'$, $y_{s2r} \gets v'/w'$
            \State $D_{s2r} \gets P_{s2r}[2]$
            \State collect $(D_{s2r}, I_{s2r}, x_{s2r}, y_{s2r})$
        \EndFor
        \State \Return $D_{s2r}, I_{s2r}, x_{s2r}, y_{s2r}$
    \end{algorithmic}
    \label{alg:1}
\end{algorithm}

\noindent\textbf{Geometric  Annotation}. 
Given the demand for a robust geometric annotator capable of handling large-scale videos, we select MegaSaM~\cite{li2025megasam} that balances both efficiency and accuracy.
We investigate open-source geometric annotation solutions, \textit{i.e.}, 
DROID-SLAM~\cite{teed2021droid},
DPVO~\cite{teed2023deep},
Fast3r~\cite{yang2025fast3r},
MonST3R~\cite{zhang2024monst3r},
and VGGT~\cite{wang2025vggt}.
In contrast to deep visual SLAM systems~\cite{teed2021droid,teed2023deep} that estimate correspondences across frames, MegaSaM is particularly effective in situations involving dynamic scenes and restricted camera parallax.
Additionally, by integrating the differentiable SLAM system with the intermediate predictions of dynamic scenes, MegaSaM outperforms 3D reconstruction schemes~\cite{teed2021droid,teed2023deep} that utilize point cloud representations from DuST3~\cite{wang2024dust3r}.
Moreover, while VGGT provides faster inference speed, MegaSAM delivers more robust performance when feature points are scarce.

\begin{figure*}[t]
  \centering
  \begin{minipage}[b]{0.24\textwidth}
    \centering
    \includegraphics[width=\textwidth]{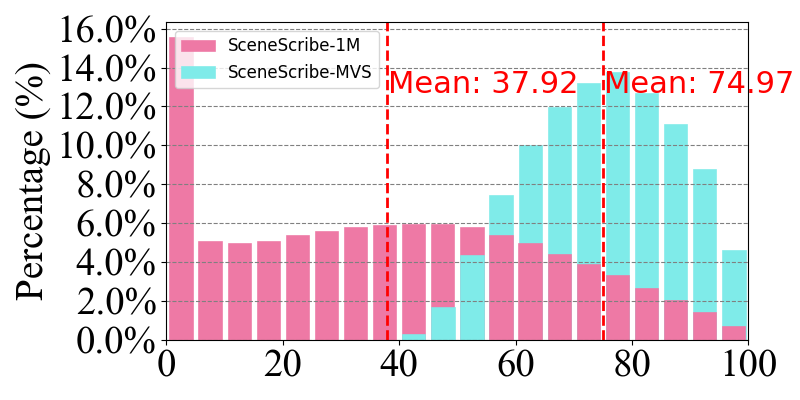}
    \centerline{(a) $s_1$ score}
    \label{fig:left3}
  \end{minipage}%
  \hspace{0.03\textwidth}
\begin{minipage}[b]{0.24\textwidth}
    \centering
    \includegraphics[width=\textwidth]{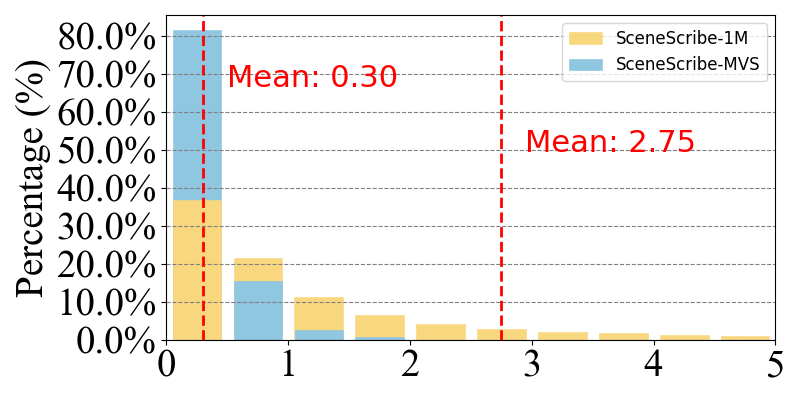}
    \centerline{(b) $s_2$ score}
    \label{fig:right1}
  \end{minipage}
    \hspace{0.05\textwidth}
  \begin{minipage}[b]{0.31\textwidth}
    \centering
    \includegraphics[width=\textwidth]{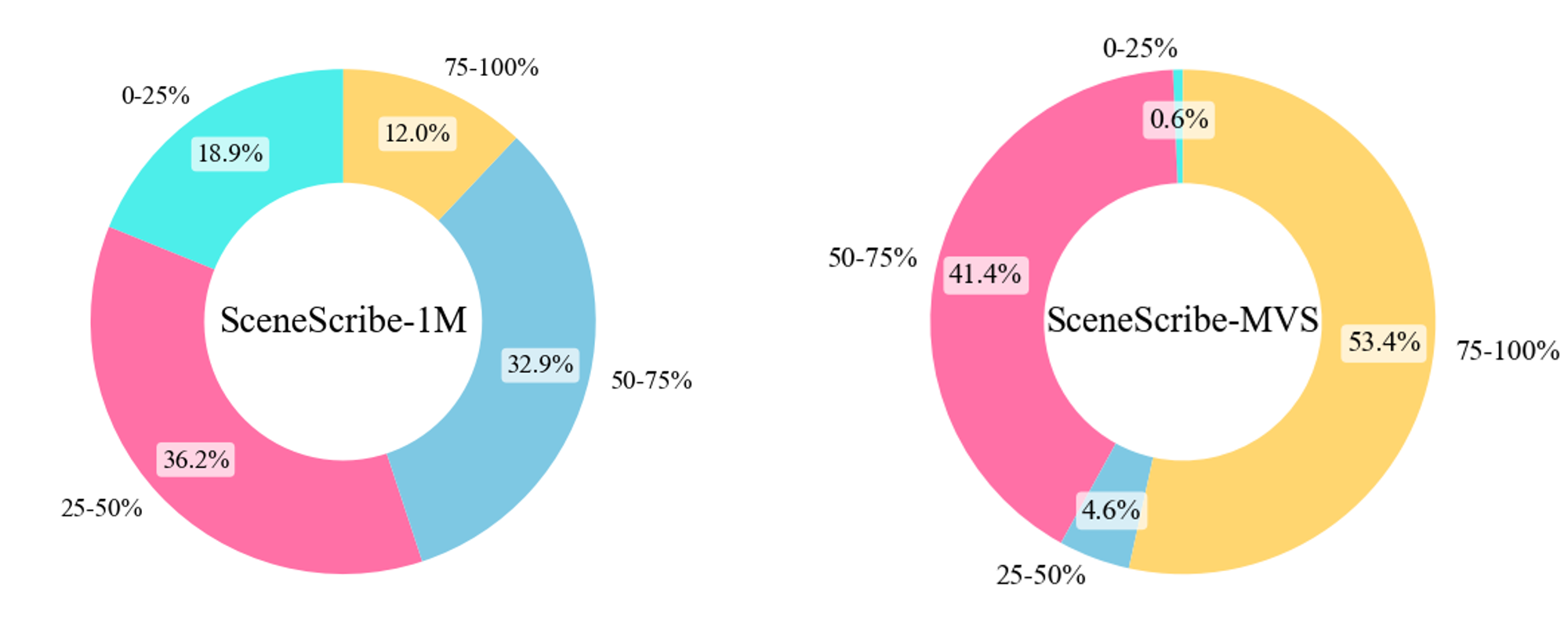}
    \centerline{(c) Visibility Radio of Tracks}
    \label{fig:right2}
  \end{minipage}
  \vspace{-0.55cm}
  \caption{\textbf{Statistics of Object Motion Metrics}. It can be observed that both object motion metrics in SceneScribe-MVS after applying the sampling strategy exhibit a greater static degree than the thresholds. This demonstrates that our sampling not only facilitates effective dynamic mask generation within SceneScribe-1M, but also improves control over the proportion of dynamics.}
    \vspace{-0.4cm}
  \label{fig:mvs}
\end{figure*}

\begin{figure*}[t]
  \centering
  \begin{minipage}[b]{0.24\textwidth}
    \centering
    \includegraphics[width=\textwidth]{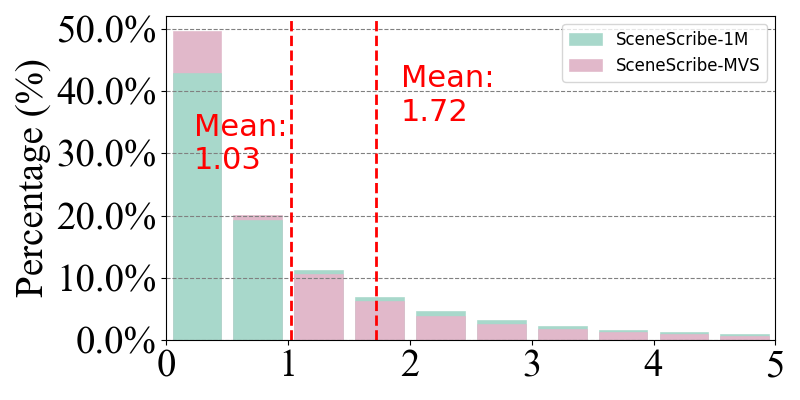}
    \centerline{(a) Distance}
    \label{fig:left4}
  \end{minipage}%
  \hspace{0.03\textwidth}
\begin{minipage}[b]{0.24\textwidth}
    \centering
    \includegraphics[width=\textwidth]{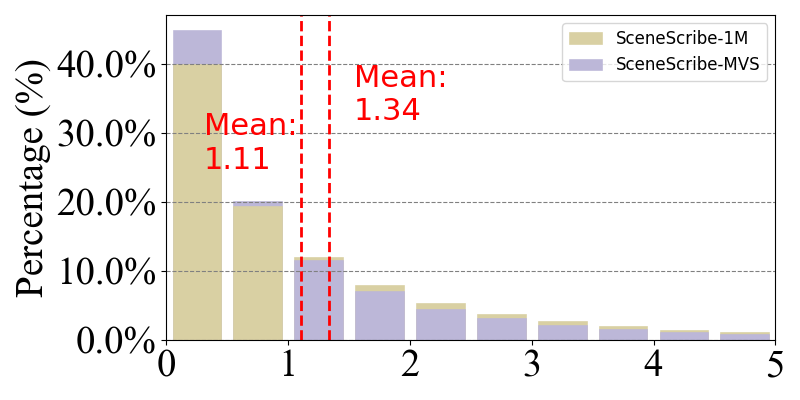}
    \centerline{(b) Rotation}
    \label{fig:right5}
  \end{minipage}
    \hspace{0.05\textwidth}
  \begin{minipage}[b]{0.31\textwidth}
    \centering
    \includegraphics[width=\textwidth]{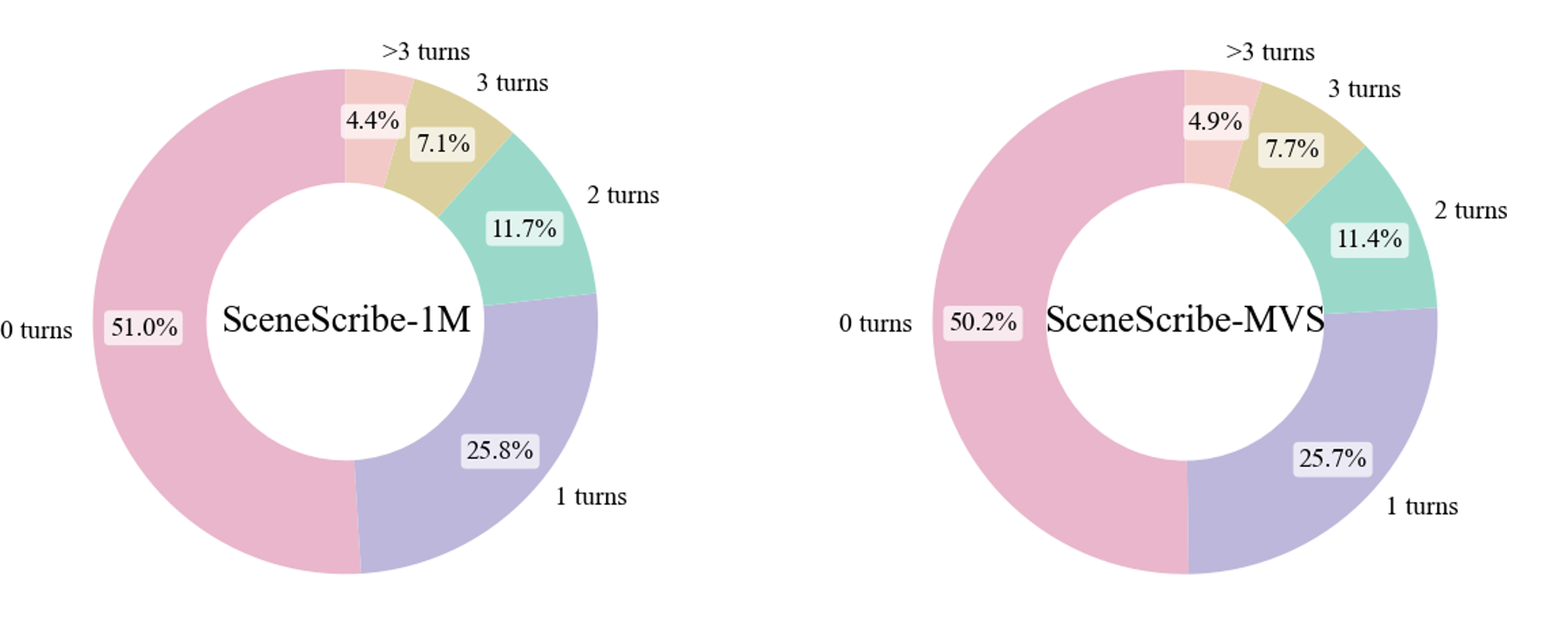}
    \centerline{(c) Turn Counts}
    \label{fig:right6}
  \end{minipage}
    \vspace{-0.55cm}
  \caption{\textbf{Statistics of Camera Motion Metrics}. The similar distributions of camera motion metrics in SceneScribe-1M and SceneScribe-MVS indicate that we disentangle camera and object motion, enabling control over object dynamics while preserving camera diversity.}
      \vspace{-0.45cm}
  \label{fig:camera}
\end{figure*}

With systematic comparisons, we employ MegaSAM for geometric annotation across three distinct aspects: 
(\textbf{\romannumeral 1}) \textit{Dynamic Motion Masks}: To efficiently handle dynamic scenes involving both camera and object motion, MegaSaM first predicts an object movement probability map, which is learned jointly with optical flow and uncertainty.
(\textbf{\romannumeral 2}) \textit{Precise Camera Parameters}: 
Building upon the DROID-SLAM~\cite{teed2021droid}, MegaSaM then integrates object movement maps and priors from mono-depth estimation (\textit{i.e.}, Depth Anything~\cite{yang2024depth} and UniDepth~\cite{piccinelli2024unidepth}) into the bundle adjustment (BA) layer, allowing for fast and robust camera tracking for unconstrained dynamic scenes;
and, (\textbf{\romannumeral 3}) \textit{Consistent Depth Maps}: 
Given the estimated camera parameters, MegaSAM optimizes the initial low-resolution disparity estimates into high-resolution video depth maps that are more accurate and temporally consistent.
Overall, we modified the official MegaSaM repository to facilitate parallel inference on over $1,000$ GPUs across multiple machines, significantly boosting the efficiency and scale of annotation. Altogether, we annotated over 4191 hours of video.

\noindent\textbf{Consistent 3D Point Tracks.} While MegaSAM produces annotations suitable for depth estimation, camera pose estimation, and scene reconstruction, it does not directly support dynamic point tracking tasks. To provide more comprehensive annotations, we further generated consistent 3D point tracks by TAPIP3D~\cite{zhang2025tapip3d}. 
Utilizing the depth and camera pose estimates from MegaSaM, TAPIP3D projects 2D video features into 3D world space, effectively compensating for camera motion.
Within this camera-stabilized spatio-temporal representation, TAPIP3D produces robust long-term 3D point tracks by iteratively refining motion estimates across multiple frames.
To facilitate compatibility with 2D tracking, we further project the 3D tracks from TAPIP3D onto the image plane using camera parameters.

\subsection{Multi-View Subset Sampling}
\label{sec:s4}
SceneScribe-1M comprises over 4,191 hours of video with diverse camera and object motions. Nonetheless, highly dynamic object motion is typically incompatible with multi-view tasks that prefer static objects.
To this end, we devise a multi-view re-projection that disentangles the motion of the camera and object. 
In addition to providing object motion masks for all scenes, we devise a sampling strategy to construct a compact subset, SceneScribe-MVS, which controls dynamic object inclusion while preserving the intensity of camera motion. 
Specifically, for each \textit{reference frame} in frame sequences,  we first select its surrounding frames within a sliding window of size $N$ as \textit{source frames} to form the \textit{sliding window pairs} $F$. 
Subsequently, we evaluate geometric and photometric consistency for each pair by utilizing annotated camera parameters and continuous video depths.
The evaluation procedure consists of four key steps, as described in Algorithm~\ref{alg:1}. 
Then, we calculate geometric and photometric errors according to the reprojected results:
\begin{align}
e_{2d} &= \sqrt{(x_{s2r} - x_r)^2 + (y_{s2r} - y_r)^2} \\
e_{3d} &= \left| D_{s2r} - D_r\right| / D_r, \quad e_{rgb} = \left\|\, I_{s2r} -  I_{r} \, \right\|_2 
\end{align}

The above errors measure the labeling consistency. Consequently, we define the motion mask by applying thresholds to filter out points exhibiting excessive errors:
\begin{align}
M_{motion} &= ({e_{2d} < \tau_{1}}) \land (e_{3d} < \tau_{2}) \land  ({e_{rgb} < \tau_{3}})
\end{align}
where $\tau_1$, $\tau_2$, and $\tau_3$ denote the thresholds. 
Based on the object motion mask $M_{motion}$ that determines the accurately annotated and static areas, we assess each scene with a score $s_1$ obtained by aggregating the mask values. Moreover, by leveraging the dynamic tracks provided by SceneScribe-1M, we calculate the average motion distance of visible points in each scene, which serves as an additional score $s_2$ for object motion intensity. Given these scores, we sample SceneScribe-MVS with thresholds $\tau_4$ and $\tau_5$. The statistics of the full set and subset are shown in Figures~\ref{fig:mvs}. The results indicate that the two scores reinforce each other, thereby substantiating the rationality of the definitions.

Additionally, we investigate the diversity of camera motion from three distinct perspectives: 
(\textbf{\romannumeral 1}) \textit{Distance} of camera trajectory;
(\textbf{\romannumeral 2}) \textit{Rotation} cumulation in camera viewing direction;
and, (\textbf{\romannumeral 3}) \textit{Turns} in camera trajectory, which counts local extrema in the sequence of angles between each frame and the start-end reference line. In Figure~\ref{fig:camera}, we present the statistics of these camera metrics. Notably, the distribution of the SceneScribe-MVS closely resembles that of the original dataset, confirming the effectiveness of the sampling strategy in disentangling camera and object motion.







\section{Experiments}

\begin{figure*}
  \centering
    \includegraphics[width=\linewidth]{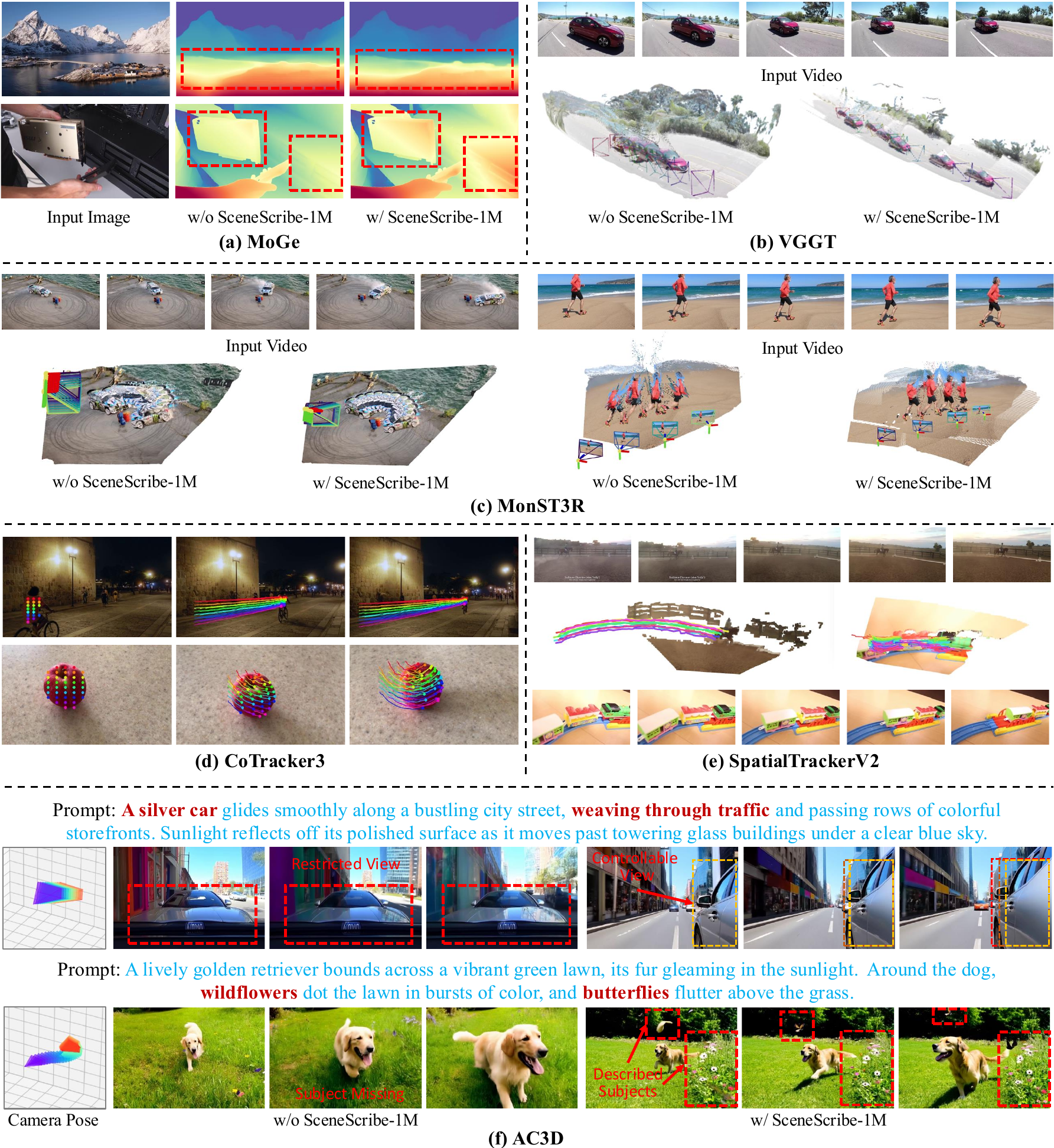}
  \vspace{-0.6cm}
  \caption{\textbf{Visualization Results of Downstream Tasks.} We conduct various downstream task on SceneScribe-1M, \textit{i.e.}, MoGe~\cite{wang2025moge} (monocular depth estimation), VGGT~\cite{wang2025vggt} (3D reconstruction), MonST3R~\cite{zhang2024monst3r} (4D reconstruction), CoTracker3~\cite{karaev2025cotracker3} (2D Point Tracking), SpatialTrackerV2~\cite{xiao2025spatialtrackerv2} (3D Point Tracking) and A3CD~\cite{bahmani2025ac3d}. These results highlight the robust applicability of SceneScribe-1M in 3D perception and video generation, offering a unified resource that effectively supports both domains at scale.}
  \vspace{-0.5cm}
  \label{fig:vis}
\end{figure*}

\begin{table*}[htbp]
  \centering
  \caption{\textbf{Evaluation of Monocular Depth Estimation on Representative Benchmarks}.}
    \vspace{-0.25cm}
  \resizebox{\linewidth}{!}{
    \begin{tabular}{lcccccccccccccccccc}
    \toprule
    \multicolumn{1}{c}{\multirow{2}[4]{*}{\textbf{Method}}} & \multicolumn{2}{c}{\textbf{NYUv2}~\cite{silberman2012indoor}} & \multicolumn{2}{c}{\textbf{KITTI}~\cite{uhrig2017sparsity}} & \multicolumn{2}{c}{\textbf{ETH3D}~\cite{schops2019bad}} & \multicolumn{2}{c}{\textbf{iBims-1}~\cite{koch2020comparison}} & \multicolumn{2}{c}{\textbf{GSO}~\cite{downs2022google}} & \multicolumn{2}{c}{\textbf{Sintel}~\cite{butler2012naturalistic}} & \multicolumn{2}{c}{\textbf{DDAD}~\cite{guizilini20203d}} & \multicolumn{2}{c}{\textbf{DIODE}~\cite{vasiljevic2019diode}} & \multicolumn{2}{c}{\textbf{Average}} \\
\cmidrule{2-19}          & Rel $\downarrow$     & $\delta_1 \uparrow$     & Rel $\downarrow$     & $\delta_1 \uparrow$     & Rel $\downarrow$     & $\delta_1 \uparrow$     & Rel $\downarrow$     & $\delta_1 \uparrow$     & Rel $\downarrow$     & $\delta_1 \uparrow$    & Rel $\downarrow$    & $\delta_1 \uparrow$    & Rel $\downarrow$    & $\delta_1 \uparrow$    & Rel $\downarrow$    & $\delta_1 \uparrow$    & Rel $\downarrow$    & $\delta_1 \uparrow$ \\
    \midrule
    \multicolumn{19}{c}{{Scale-invariant depth map}} \\
    \midrule
    Moge (w/o SceneScribe) & 3.44  & \textbf{98.4}  & 4.25  & 97.8  & \textbf{3.36}  & \textbf{98.9}  & 3.46  & 97.0    & \textbf{1.47}  & \textbf{100}   & \textbf{19.3}  & \textbf{73.4}  & 9.17  & 90.5  & 4.89  & 94.7  & 6.17 & 93.8 \\
    Moge (w SceneScribe-1M) & \textbf{3.42}  & 98.3  & \textbf{4.13}  & \textbf{97.9}  & 3.45  & 98.7  & \textbf{3.26}  & \textbf{98.0}    & \textbf{1.47}  & \textbf{100}   & 19.6  & 72.0    & \textbf{8.95}  & \textbf{91.5}  & \textbf{4.82}  & \textbf{95.3}  & \textbf{6.14} & \textbf{94.0} \\
    \midrule
    \multicolumn{19}{c}{{Affine-invariant depth map}} \\
    \midrule
    Moge (w/o SceneScribe) & 2.92  & \textbf{98.6}  & 3.94  & 98.0    & \textbf{2.69}  & \textbf{99.2}  & 2.74  & 97.9  & \textbf{0.94}  & \textbf{100}   & \textbf{13.0}    & \textbf{83.2}  & 8.40   & 92.1  & 3.16  & \textbf{97.5} & 4.72 & 95.8 \\
    Moge (w SceneScribe) & \textbf{2.83}  & \textbf{98.6}  & \textbf{3.80}   & \textbf{98.1}  & 2.78  & \textbf{99.2}  & \textbf{2.46}  & \textbf{98.5}  & 0.95  & \textbf{100}   & 13.2  & 82.7  & \textbf{8.31}  & \textbf{92.4}  & \textbf{3.14}  & \textbf{97.5}  & \textbf{4.68} & \textbf{95.9} \\
    \midrule
    \multicolumn{19}{c}{{Affine-invariant disparity map}} \\
    \midrule
    Moge (w/o SceneScribe) & 3.38  & 98.6  & 4.05  & \textbf{98.1}  & \textbf{3.11}  & \textbf{98.9}  & 3.23  & 98.0    & \textbf{0.96}  & \textbf{100}   & 18.4  & 79.5  & 8.99  & 91.5  & \textbf{3.98}  & \textbf{97.2}  & 5.76 & 95.2 \\
    Moge (w SceneScribe) & \textbf{3.35}  & \textbf{98.7}  & \textbf{3.99}  & \textbf{98.1}  & 3.19  & \textbf{98.9}  & \textbf{2.97}  & \textbf{98.4}  & \textbf{0.96}  & \textbf{100}   & \textbf{18.2}  & 79.4  & \textbf{8.74}  & \textbf{91.9}  & 4.01  & \textbf{97.2}  & \textbf{5.68} & \textbf{95.3} \\
    \bottomrule
    \end{tabular}}%
  \label{tab:moge}%
     \vspace{-0.25cm}
\end{table*}%

\begin{table*}[!t]
    \centering
    \caption{\textbf{Evaluation of Scene Reconstruction on Representative Benchmarks}.}
       \vspace{-0.25cm}
    \begin{minipage}[t]{.49\textwidth}
        \centering
        \centerline{(a) 3D Reconstruction  on CO3Dv2~\cite{reizenstein2021common} and ETH3D~\cite{butler2012naturalistic}.}
        \resizebox{\textwidth}{!}{
            \setlength{\tabcolsep}{1.2mm}{
    \begin{tabular}{lccccc}
    \toprule
          & \multicolumn{2}{c}{\textbf{Pose Estimation}} & \multicolumn{3}{c}{\textbf{Point Map Estimation}} \\
    \midrule
    \textbf{Method} & {AUC$_{30}$} $\uparrow$ & {AUC$_{15}$} $\uparrow$ & \multicolumn{1}{c}{ACC.} $\downarrow$ & Comp. $\downarrow$ & Overall $\downarrow$ \\
    \midrule
    VGGT (w/o SceneScribe-1M) & 89.5  & 83.4   & \textbf{{0.873}} & \textbf{0.482} & \textbf{0.677} \\
    VGGT (w SceneScribe-1M) &  \textbf{89.9}   &  \textbf{83.8}   &   0.890 &0.504&	0.697\\
    \bottomrule
    \end{tabular}%
            }
        }
        \end{minipage}
        \hfill
        \begin{minipage}[t]{.49\textwidth}
        \centering
        \centerline{(b) 4D Reconstruction  on Sintel~\cite{butler2012naturalistic} Dataset.}
        \resizebox{\textwidth}{!}{
            \setlength{\tabcolsep}{1.0mm}{
                \begin{tabular}{lccccc}
                \toprule
                \multicolumn{1}{c}{\multirow{2}[4]{*}{\textbf{Method}}} & \multicolumn{3}{c}{\textbf{Pose Estimation}} & \multicolumn{2}{c}{\textbf{Depth Estimation}} \\
            \cmidrule{2-6}          & \multicolumn{1}{c}{ATE $\downarrow$} & \multicolumn{1}{c}{RPE trans $\downarrow$} & \multicolumn{1}{c}{RPE rot $\downarrow$} & {Rel $\downarrow$} & {$\delta_1 \uparrow$} \\
                \midrule
                MonST3R (w/o SceneScribe) & 0.108 & 0.042 & 0.732 & 0.335 & \textbf{58.5} \\
                MonST3R  (w SceneScribe) & \textbf{0.099} & \textbf{0.038} & \textbf{0.685} &  \textbf{0.320} & 58.1 \\
                \bottomrule
                \end{tabular}
            }
        }
    \end{minipage}
     \vspace{-0.25cm}
    \label{tab::recs}
\end{table*}

\begin{table*}[!t]
    \centering
    \caption{\textbf{Evaluation of Dynamic Point Tracking on Representative Benchmarks.}}
       \vspace{-0.25cm}
    \begin{minipage}[t]{.49\textwidth}
        \centering
        \centerline{(a) 2D Point Tracking on TAP-Vid~\cite{doersch2022tap} benchmarks.}
        \resizebox{\textwidth}{!}{
            \setlength{\tabcolsep}{1.0mm}{
                \begin{tabular}{lcccccccccc}
                \toprule
                      & \multicolumn{3}{c}{\textbf{Kinetics}} & \multicolumn{3}{c}{\textbf{RGB-S}} & \multicolumn{3}{c}{\textbf{DAVIS}} & \multicolumn{1}{l}{\textbf{Mean}} \\
                \midrule
                \textbf{Method} & AJ $\uparrow$    & $\delta_{avg}^{vis}$ $\uparrow$     & OA $\uparrow$    & AJ $\uparrow$     & $\delta_{avg}^{vis}$ $\uparrow$    & OA  $\uparrow$   & AJ $\uparrow$    & $\delta_{avg}^{vis}$  $\uparrow$   & OA $\uparrow$    & $\delta_{avg}^{vis}$ $\uparrow$ \\
                \midrule
                CoTracker3 (w/o SceneScribe) & 54.7  & 67.8  & 87.4  & 74.3  & 85.2  & 92.4  & 64.4  & 76.9  & 91.2  & 76.6 \\
                CoTracker3 (w SceneScribe) & \textbf{55.5} & \textbf{68.4} & \textbf{88.2} & \textbf{74.9} & \textbf{86.3} & \textbf{92.8} & \textbf{64.5} & \textbf{77.6} & \textbf{92.0} & \textbf{77.4} \\
                \bottomrule
                \end{tabular}
            }
        }
        \end{minipage}
        \hfill
        \begin{minipage}[t]{.49\textwidth}
        \centering
        \centerline{(b) 3D Point Tracking on TAPVid-3D~\cite{koppula2024tapvid} benchmarks}
        \resizebox{\textwidth}{!}{
            \setlength{\tabcolsep}{1.0mm}{
                \begin{tabular}{lccccccccc}
                \toprule
                      & \multicolumn{3}{c}{\textbf{Aria}} & \multicolumn{3}{c}{\textbf{Pstudio}} & \multicolumn{3}{c}{\textbf{Average}} \\
                \midrule
                \textbf{Method} & AJ $\uparrow$    & APD $\uparrow$  & OA    & AJ $\uparrow$    & APD $\uparrow$   & OA $\uparrow$   & AJ $\uparrow$    & APD $\uparrow$   & OA $\uparrow$ \\
                \midrule
                SpatialTrackerV2(w/o SceneScribe) & 24.6  & 34.7  & 93.6  & 21.9  & 32.1  & 87.4  & 23.25 & 33.4  & 60.3 \\
                SpatialTrackerV2 (w SceneScribe-1M) & \textbf{24.7} & \textbf{34.7} & \textbf{93.8} & \textbf{22.3} & \textbf{32.5} & \textbf{87.9} & \textbf{23.5} & \textbf{33.6} & \textbf{60.6} \\
                \bottomrule
                \end{tabular}
            }
        }
    \end{minipage}
       \vspace{-0.3cm}
    \label{tab::track}
\end{table*}

\begin{table}[!t]
  \centering
  \caption{\textbf{Text/Pose-to-Video Evaluation on RealEstate10K~\cite{zhou2018stereo}}.}
     \vspace{-0.2cm}
  \resizebox{\linewidth}{!}{
    \begin{tabular}{lccccc}
    \toprule
    \textbf{Method} & \multicolumn{1}{c}{TransErr $\downarrow$} & \multicolumn{1}{c}{RotErr $\downarrow$} & \multicolumn{1}{c}{FID $\downarrow$} & \multicolumn{1}{c}{FVD $\downarrow$} & \multicolumn{1}{c}{CLIP $\uparrow$} \\
    \midrule
    AC3D (w/o SceneScribe-1M) & 0.374 & 0.039 & 1.27  & 38.20  & 28.62 \\
    AC3D (w SceneScribe-1M) & \textbf{0.318} & \textbf{0.026} & \textbf{1.19}  & \textbf{35.15} & \textbf{29.98} \\
    \bottomrule
    \end{tabular}}
  \label{tab:ac3d}%
   \vspace{-0.6cm}
\end{table}%

\subsection{Implementation Details}
 For the curation pipeline, 
 we parallelized the inference of MegaSaM and TAPIP3D using batch processing and multithreading.
 We utilize more than $1,000$ NVIDIA H20 GPUs across multiple machines. The overall annotation process consumed about $150k$ GPU hours.
Unless otherwise specified, all downstream models follow the original training configurations, including hyperparameters and the number of GPUs. To ensure a fair comparison, all baselines are evaluated under their officially specified configurations.


\subsection{Downstream Tasks}
To comprehensively evaluate the reliability and applicability of the annotation pipeline, we conduct multiple downstream tasks on the SceneScribe-1M, including monocular depth estimation~\cite{wang2025moge}, Scene reconstruction~\cite{wang2025vggt,zhang2024monst3r}, dynamic point tracking~\cite{karaev2025cotracker3,xiao2025spatialtrackerv2}, and generative tasks~\cite{bahmani2025ac3d}. The qualitative results are illustrated in Figure~\ref{fig:vis}.

\noindent\textbf{Monocular Depth Estimation}. MagaSaM optimizes continuous video depth by leveraging temporal information, making the per-frame depth maps suitable for monocular depth estimation tasks. 
Accordingly, we retrain MoGe~\cite{wang2025moge} by integrating the SceneScribe with the original TartanAir~\cite{wang2020tartanair} datasets.
Notably, as the TartanAir dataset is synthetic, it inherently provides high-quality annotations. Thus, the improvements achieved by integrating SceneScribe-1M (as shown in Figure~\ref{fig:vis} (a) and Table~\ref{tab:moge}) demonstrate the effectiveness of our annotation pipeline.

\noindent\textbf{Scene Reconstruction}.
Since SceneScribe-1M provides annotations for continuous video depth and camera pose, it can be directly applied to the 3D reconstruction of VGGT~\cite{wang2025vggt} and 4D reconstruction of MonST3R~\cite{zhang2024monst3r}.
As shown in Table~\ref{tab::recs} (a), we begin by assessing the impact of SceneScribe-1M on the 3D reconstruction performance of VGGT. The quantitative results indicate that SceneScribe-1M facilitates camera pose estimation, while slightly compromising the performance of point map estimation, consistent with the qualitative results in Figure~\ref{fig:vis} (b).
In Table~\ref{tab::recs} (b), we evaluate 4D reconstruction capabilities on the Sintel dataset to assess model performance under diverse dynamic scene conditions. SceneScribe further improves the camera pose estimation capability of MonST3R, while preserving its strength in depth estimation. In addition,  we provide a visualization of the 4D reconstruction in Figure~\ref{fig:vis} (c).

\noindent\textbf{Dynamic Point Tracking}.
SceneScribe-1M contains point tracks annotated by TAPIP3D~\cite{zhang2025tapip3d} based on the geometric format of MegaSAM~\cite{li2025megasam}, which makes it suitable for CoTracker3~\cite{karaev2025cotracker3} (2D Point Tracking) and SpatialTrackerV2~\cite{xiao2025spatialtrackerv2} (3D Point Tracking).
As shown in Tables~\ref{tab::track}, the results on TAP-Vid and TAPVid-3D benchmarks demonstrate that SceneScribe-1M achieves annotation accuracy comparable to that of standard datasets such as Kubric~\cite{greff2022kubric}, PointOdyssey~\cite{zheng2023pointodyssey}, and Dynamic Replica~\cite{karaev2023dynamicstereo}.
Meanwhile, the large-scale annotation further guarantees the generalizability of dynamic point tracking, as demonstrated by the visualizations in Figures~\ref{fig:vis} (d) and~\ref{fig:vis} (e).


\noindent\textbf{Text/Pose-to-Video Generation}.
Given the textual descriptions and camera pose annotations provided in SceneScribe-1M, we utilize the AC3D~\cite{bahmani2025ac3d} model to demonstrate the feasibility of the text/pose-to-video task. Compared to RealEstate10K~\cite{zhou2018stereo}, the larger SceneScribe-1M provides superior diversity in video content and increased precision in camera pose annotations. These advantages lead to improved generation quality and camera controllability, as shown in the qualitative results in Figure~\ref{fig:vis} (f) and the quantitative results in Table~\ref{tab:ac3d}, respectively.

\section{Conclution}

In this work, we address the pressing need for large-scale datasets that jointly advance 3D geometric perception and video synthesis. By introducing SceneScribe-1M, a multi-modal, large-scale video dataset comprehensively annotated with detailed semantics and 3D information, we bridge an important gap between these two domains. Various benchmarks demonstrate that SceneScribe-1M supports a wide range of downstream tasks, including depth estimation, scene reconstruction, dynamic point tracking, and camera-controlled text-to-video generation. By making SceneScribe-1M openly available, we aim to facilitate broader research progress and provide a unified resource for developing world foundation models capable of generating semantic-rich and physically grounded video content.
\newpage

{
    \small
    \bibliographystyle{ieeenat_fullname}
    \bibliography{main}
}


\end{document}